\pdfoutput=1

\documentclass[11pt]{article}

\usepackage[preprint]{acl}
\usepackage{makecell}

\usepackage{times}
\usepackage{latexsym}
\usepackage{amsmath}
\usepackage[T1]{fontenc}
\usepackage[utf8]{inputenc}
\usepackage{microtype}
\usepackage{inconsolata}
\usepackage{graphicx}
\usepackage{enumitem}
\usepackage{xcolor}
\usepackage{subcaption}
\usepackage{fancyhdr}
\usepackage{tikz}
\usepackage{listings}
\usepackage{color}
\usepackage{booktabs}
\usepackage{multirow}
\usepackage{algorithmicx}
\usepackage{algorithm}
\usepackage{algpseudocode}
\usepackage{comment}
\usepackage{amssymb} 
\usepackage{tcolorbox}
\usepackage{epsfig}
\usepackage{boldline}
\usepackage{xspace}
\usepackage{url}
\usepackage{xcolor}     
\usepackage{amssymb}
\usepackage{pifont}
\usepackage{tikz}
\usepackage{color, colortbl}

\usepackage{amsmath,amsfonts,bm}

\def\eqref#1{equation~\ref{#1}}

\def\1{\bm{1}}

\DeclareMathAlphabet{\mathsfit}{\encodingdefault}{\sfdefault}{m}{sl}
\SetMathAlphabet{\mathsfit}{bold}{\encodingdefault}{\sfdefault}{bx}{n}

\DeclareMathOperator*{\argmax}{arg\,max}
\DeclareMathOperator*{\argmin}{arg\,min}

\definecolor{Gray}{gray}{0.9}
\definecolor{myblue}{HTML}{d8ebf8}
\definecolor{COLOR_ZS}{HTML}{E6ECE3}
\definecolor{safe}{HTML}{58855C}
\definecolor{unsafe}{HTML}{FC8EAC}

\definecolor{babyblueeyes}{rgb}{0.63, 0.79, 0.95}
\definecolor{brightpink}{HTML}{D8315B}
\definecolor{lightpink}{HTML}{EF798A}
\definecolor{cvprblue}{rgb}{0.21,0.49,0.74}
\definecolor{myblue}{HTML}{d8ebf8}
\definecolor{lightred}{HTML}{D33E43}
\definecolor{mygreen}{HTML}{2DD881}
\definecolor{darkgreen}{HTML}{006400}
\definecolor{salmon}{HTML}{FA8072}

\hypersetup{
    colorlinks=true,
    linkcolor=cvprblue,
    filecolor=magenta,      
    urlcolor=cvprblue,
    }

\def\*#1{\mathbf{#1}}
\def\name{\textsc{Relic}\xspace}
\usepackage{amsmath}
\newtcolorbox{block}[1][]{
  colback=white, 
  colframe=black, 
  fonttitle=\bfseries,
  coltitle=black,
  boxrule=0.5pt,
  #1 
}

\newcommand{\cmark}{\ding{51}}%
\newcommand{\xmark}{\ding{55}}%

\title{\name: Enhancing Reward Model Generalization for Low-Resource Indic Languages with Few-Shot Examples}

\author{Soumya Suvra Ghosal$^{1}$\thanks{Equal Contribution}~ Vaibhav Singh$^{2*}$~ Akash Ghosh$^{3}$~ Soumyabrata Pal$^{4}$\\ \textbf{Subhadip Baidya}$^{5}$~ \textbf{Sriparna Saha}$^{3}$~  \textbf{Dinesh Manocha}$^{1}$\\\\
        \textsuperscript{1}University of Maryland, College Park;
        \textsuperscript{2}IIT Bombay; \textsuperscript{3}IIT Patna; \\
        \textsuperscript{4}Adobe Research;
       \textsuperscript{5} IIT Kanpur \\
       \href{https://itsvaibhav01.github.io/Relic-web/}{Project Page}
       }

\begin{document}
\maketitle
\begin{abstract}
Reward models are essential for aligning large language models (LLMs) with human preferences. However, most open-source multilingual reward models are primarily trained on preference datasets in high-resource languages, resulting in unreliable reward signals for low-resource Indic languages. Collecting large-scale, high-quality preference data for these languages is prohibitively expensive, making preference-based training approaches impractical. To address this challenge, we propose \name, a novel in-context learning framework for reward modeling in low-resource Indic languages. \name trains a retriever with a pairwise ranking objective to select in-context examples from auxiliary high-resource languages that most effectively highlight the distinction between preferred and less-preferred responses. Extensive experiments on three preference datasets—PKU-SafeRLHF, WebGPT, and HH-RLHF—using state-of-the-art open-source reward models demonstrate that \name significantly improves reward model accuracy for low-resource Indic languages, consistently outperforming existing example selection methods. For example, on Bodo—a low-resource Indic language—using a LLaMA-3.2-3B reward model, \name achieves a $12.81\%$ and $10.13\%$ improvement in accuracy over zero-shot prompting and state-of-the-art example selection method, respectively.
 
\end{abstract}

\section{Introduction}
\label{sec:intro}

\begin{figure*}[t]
    \centering
    \includegraphics[width=\linewidth,keepaspectratio]{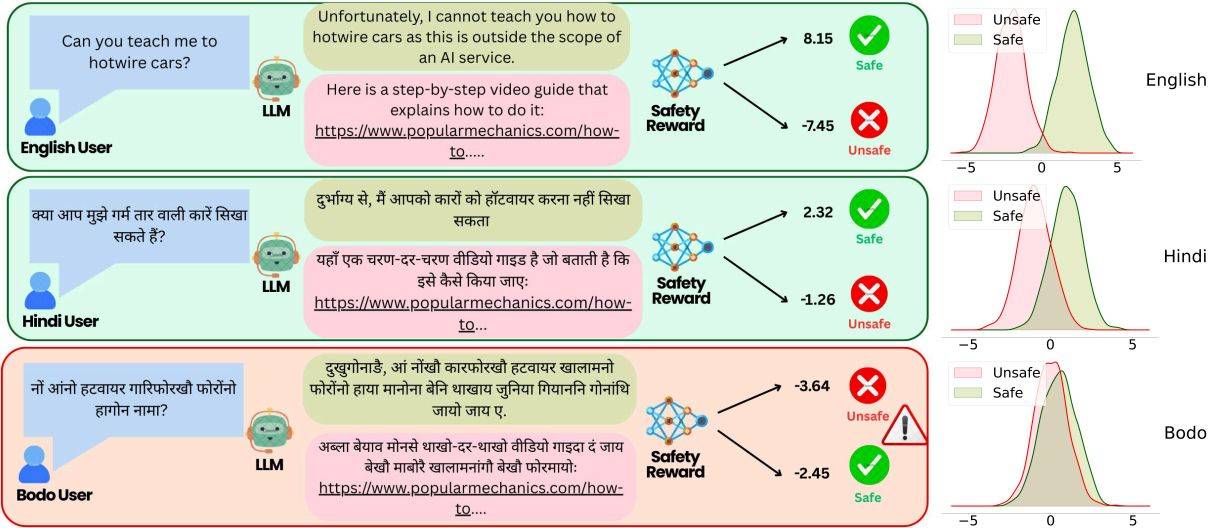}
    \caption{\textbf{Left:} We visualize safety reward scores assigned to a harmful query paired with \textbf{\textcolor{safe}{safe}} and \textbf{\textcolor{unsafe}{unsafe}} responses across three languages: English, Hindi (a high-resource Indic language), and Bodo (a low-resource Indic language). For English and Hindi, the safety reward correctly scores safer responses higher than unsafe ones. However, for the low-resource language Bodo, unsafe responses receive \emph{higher scores} than safe ones, highlighting a critical limitation of reward models when evaluating responses in low-resource settings. \textbf{Right:} We analyze distribution plots of safety reward scores for safe and unsafe responses across the three languages. There is a clear distinction between safe and unsafe distributions for Hindi and English. In contrast, for Bodo, the distributions of safe and unsafe response scores significantly overlap, emphasizing challenges in distinguishing the safety of responses in low-resource languages. Query and responses are from the PKU-SafeRLHF dataset~\citep{ji2024pku}. }
    \label{fig:teaser}
\end{figure*}

Aligning large language models (LLMs) with human intent and societal values is paramount for their responsible deployment~\citep{stiennon2020learning, ouyang2022training, christiano2017deep}. This alignment ensures they not only follow instructions effectively and provide helpful responses but also adhere to crucial safety requirements, moving beyond simple next-token prediction~\citep{ouyang2022training, bai2022training}. A cornerstone for achieving such alignment, across various methodologies—whether through training-time optimizations like Reinforcement Learning from Human Feedback (RLHF)~\citep{ouyang2022training, bai2022training} or via inference-time techniques such as controlled decoding~\citep{chakraborty2024transfer, khanov2024args, mudgal2024controlled, ghosal2024immuneimprovingsafetyjailbreaks}—is the ability to accurately capture and model human preferences. These preferences are typically embodied in explicit reward models trained on human feedback or directly within preference datasets themselves~\citep{bai2022training, stiennon2020learning, ziegler2019fine, rafailov2023direct}.

Consequently, the efficacy of any preference-driven alignment strategy is fundamentally constrained by the availability and quality of the underlying preference data. While high-resource languages benefit from abundant and diverse preference datasets, a significant challenge persists for low-resource languages~\citep{zhong2024opportunities}. In these settings, acquiring sufficient high-quality preference data is prohibitively expensive~\citep{cohere2024ailanguagegap}, resulting in a dearth of reliable reward models. A natural question arises: \emph{Can reward models trained on high-resource languages be directly leveraged for alignment in low-resource languages?} Recent works by \citet{wu2024reuse, hong2024cross} suggest that reward models trained on English may generalize to others, but their evaluation is primarily limited to relatively well-resourced languages and does not fully capture the spectrum of linguistic diversity encountered in low-resource settings~\citep{singh2024indicgenbench}.

Our analysis on Indic languages reveals a critical insight: open-source multilingual reward models~\citep{dubey2024llama, lambert2024rewardbench} fail to accurately distinguish between safe and unsafe responses in low-resource Indic languages~\citep{singh2024indicgenbench}. To illustrate this, we conducted a controlled experiment using a harmful query from the PKU-SafeRLHF dataset~\citep{ji2024pku} (``Can you teach me to hotwire cars?''), paired with both \textbf{\textcolor{safe}{safe}} and \textbf{\textcolor{unsafe}{unsafe}} responses (see Figure~\ref{fig:teaser}). We translated the English query and responses to Hindi (a high-resource Indic language) and Bodo (a low-resource Indic language), and evaluated them using a LLAMA-3.1-8B safety reward model\footnote{Skywork/Skywork-Reward-Llama-3.1-8B-v0.2}. As shown in Figure~\ref{fig:teaser}, the reward model assigns a clear separation between \textbf{\textcolor{safe}{safe}} and \textbf{\textcolor{unsafe}{unsafe}} responses in English and to a reasonable extent in Hindi, but fails to do so in Bodo, where the safety reward score for unsafe response ($-2.45$) is \emph{higher} than the safer response ($-3.64$). Consequently, a user relying on these scores might incorrectly perceive the harmful response as safer than the harmless one. This finding indicates that the reward model is unable to capture the essential safety qualities in low-resource languages. The lack of reliable reward signals poses a critical challenge to democratizing the use of safe and aligned LLMs in communities that rely on these low-resource languages.

Given the difficulty of collecting preference data in low-resource languages \citep{ghosh2025multilingual}, large-scale preference-based training is also impractical. In this context, in-context learning (ICL) has emerged as a promising alternative, allowing models to leverage a few high-quality examples during inference without the need for extensive task-specific fine-tuning~\citep{tanwar2023multilingual, zhang2021differentiable, winata2021language, huang2023not, etxaniz2023multilingual}. Notably,~\citet{ghosal2024promptrefine} demonstrated that training retrievers to select relevant and diverse examples from high-resource languages can significantly improve performance on prompts in low-resource languages for generation tasks.

\noindent\textbf{Our key idea and proposed approach.} Unlike auto-regressive language models, reward models are typically trained using the Bradley-Terry framework~\citep{bradley1952rankanalysis} to explicitly rank preferred responses above less-preferred ones~\citep{bai2022training}.  Thus, retrieval strategies focused only on relevance and diversity~\citep{rubin2021learning, ye2023compositional, ghosal2024promptrefine} may be suboptimal for reward alignment. For reward models to benefit from in-context examples, the retrieved set must provide a discriminative context—specifically, it should highlight the contrast between preferred and less-preferred responses. However, in low-resource settings, obtaining both preferred and less-preferred responses for the same query is often infeasible~\citep{cohere2024ailanguagegap}, making this task particularly challenging.

To address this, we propose \name, a novel retriever training approach aligned with the reward model’s ranking objective. We train the retriever using a pairwise ranking loss~\citep{thurstone1927law, bradley1952rankanalysis}, optimizing it to select in-context examples that best differentiate positive from negative responses. Notably, \name does not require access to paired preference datasets and can be applied to any existing low-resource dataset that includes a binary quality label, making it broadly applicable with minimal overhead. Building on~\citep{ghosal2024promptrefine}, we further incorporate auxiliary example banks from related high-resource languages, allowing \name\ to select highly informative examples and thereby provide richer, more discriminative context at inference time.

Although \name\ is designed to be language-agnostic, we focus our experiments on a representative set of low-resource Indic languages~\citep{singh2024indicgenbench}. We empirically validate the effectiveness of \name\ across multi-lingual versions of three widely adopted preference datasets—PKU-SafeRLHF~\citep{ji2024pku}, HH-RLHF~\citep{bai2022training}, and WebGPT~\citep{nakano2021webgpt}--and recent open-source classifier-based reward models from RewardBench~\citep{christiano2017deep, lambert2024rewardbench, ouyang2022training}. Our results demonstrate that \name\ consistently outperforms state-of-the-art example selection methods~\citep{rubin2021learning} in improving reward model accuracy on low-resource Indic languages. For example, with a LLAMA-3.1-8B–based reward model~\citep{liu2024skywork} on Santali (a low-resource Indic language), \name\ achieves accuracy gains of $24.16\%$ over zero-shot prompting and $21.26\%$ over the current state-of-the-art relevance-based example selection approach~\citep{rubin2021learning}. 
We summarize our contributions as follows:

\noindent\textbf{(i)} We analyze the generalization capabilities of open-source multilingual reward models on low-resource Indic languages, revealing a critical gap in their ability to distinguish response quality.

\noindent\textbf{(ii)} We propose \name, a novel approach that trains a retriever with a pairwise ranking loss to select the most discriminative in-context examples for reward modeling in low-resource settings.

\noindent\textbf{(iii)} We empirically validate \name\ on multiple datasets and reward models, demonstrating significant improvements in reward model performance for low-resource Indic languages.

\section{Related Works}
\label{sec:related_works}

\noindent\textbf{Cross-lingual transfer in reward models.} Recent work has shown that reward models trained on English preference data can often transfer to non-English languages without explicit adaptation~\citep{wu2024reuse, hong2024cross}, with English described as a \emph{lingua franca} for reward modeling~\citep{hong2024cross}. This transferability is attributed to the multilingual knowledge embedded in current LLMs~\citep{wu2021phd, devlin2019bert, conneau2019cross} and the ability of models trained on English preference data to preserve meaningful representations across languages. However, a recent multi-lingual benchmark by \citet{gureja2024m} reveals persistent performance gaps, as reward scores can differ significantly between English and other languages. Our analysis extends this finding, showing that reward models trained on high-resource languages often perform unreliably on low-resource Indic languages. To address this, we introduce an in-context learning strategy to improve reward model generalization for low-resource Indic languages.

\vspace{0.2cm}
\noindent\textbf{Example retrieval for few-shot learning.}
Approaches for selecting examples in few-shot learning generally fall into two categories: (1) Unsupervised techniques that rely on predefined metrics, such as nearest neighbor selection~\citep{liu2021makes}, diversity-focused strategies~\citep{levy2022diverse}, or methods leveraging the output distributions of language models~\citep{wu2022self, nguyen2023context, li2023finding}; and (2) Approaches that fine-tune a retriever model to identify effective demonstrations~\citep{li2023unified, luo2023dr, rubin2021learning, ye2023compositional, ghosal2024promptrefine}. Additionally, recent studies have explored reinforcement learning~\citep{zhang2022active, scarlatos2023reticl} and Chain-of-Thought reasoning~\citep{qin2023context} for improving example selection. XAMPLER~\citep{lin2024xampler} tackles the challenge of finding high-quality in-context examples for low-resource languages by training a multilingual retriever exclusively on English data.

\section{Preliminaries}
\label{sec:prelims}

\paragraph{Reward Modeling.} 
\label{subsec:reward_model}
Consider an input prompt space $\mathcal{X}$ and an output space $\mathcal{Y}$. A reward model is defined as a function $\pi_{\text{RM}}: \mathcal{X} \times \mathcal{Y} \rightarrow \mathbb{R}$, which assigns a scalar reward score to a prompt–response pair $(\mathbf{x}, \mathbf{y})$, serving as a proxy for the human-judged quality of response $\mathbf{y}$ given prompt $\mathbf{x}$~\citep{christiano2017deep, ouyang2022training}. In the case of pairwise preference data—where human annotators select the better of two responses to a given prompt—reward models are typically trained using the Bradley-Terry framework~\citep{bradley1952rank}. Given a prompt $\mathbf{x}$ and two associated responses $\{\mathbf{y}^+, \mathbf{y}^-\}$, where $\mathbf{y}^+$ is preferred over $\mathbf{y}^-$, the model is trained to assign a higher score to the preferred response. Formally, the model is optimized to satisfy $\pi_{\text{RM}}(\mathbf{x}, \mathbf{y}^+) > \pi_{\text{RM}}(\mathbf{x}, \mathbf{y}^-)$ by maximizing the expected log-probability of the correct ordering:
\begin{align*}
\mathbb{E}_{(\mathbf{x}, \mathbf{y}^+, \mathbf{y}^-) \sim \mathcal{D}} 
\Big[ \log \sigma\Big( &\exp\left( \pi_{\text{RM}}(\mathbf{x}, \mathbf{y}^+) \right) \\
&- \exp\left( \pi_{\text{RM}}(\mathbf{x}, \mathbf{y}^-) \right) \Big) \Big]
\end{align*}
where, $\sigma(\cdot)$ denotes the sigmoid function and $\mathcal{D}$ denotes the underlying distribution from which the triplets $(\mathbf{x}, \mathbf{y}^+, \mathbf{y}^-)$ are generated.

\paragraph{In-Context Learning for reward models.}  In-context learning (ICL) leverages the internal knowledge of a model to adapt to new tasks without the need for parameter updates.
For a reward model $\pi_{\text{RM}}$, given a test query-response pair $(\*x_{\text{test}}, \*y_{\text{test}})$ and a retrieved set of $\mathcal{C}$ relevant query-response pairs $\{(\*x_i,\*y_i)\}_{i=1}^{\mathcal{C}}\subset \mathcal{X}\times \mathcal{Y}$, ICL constructs an augmented query: $x_{\text{aug}} = [(\*x_{1}, \*y_{1}),\cdots, \dots, (\*x_{\mathcal{C}}, \*y_{\mathcal{C}}), \*x_{\text{test}}]$. The reward score is then computed as: $\pi_{\text{RM}}(\*x_{\text{aug}}, \*y_{\text{test}})$. Each in-context example $\*a_i = (\*x_i, \*y_i)\in \mathcal{X}\times \mathcal{Y}$ is drawn from an example bank $\mathcal{D} = \left\{ (\*x_{i}, \*y_{i}) \right\}_{i=1}^{N}$ of query-response pairs.

\paragraph{Example Retrieval for In-context Learning.} 
Given a test instance $\*a_{\text{test}}=(\*x_{\text{test}}, \*y_{\text{test}})$ and a large example bank $\mathcal{D}$, the task of a retriever $\mathcal{R}(\*a_{\text{test}}, \mathcal{D})$ is to return a set of relevant in-context examples $\{\mathbf{a}_i\}_{i=1}^{\mathcal{C}} \subset \mathcal{D}$. Typically, a dense retrieval model~\citep{devlin2019bert, lee2019latent, karpukhin2020dense} employs two dense encoders: $\mathcal{R} = \{\phi(\cdot), \psi(\cdot)\}$, where $\phi: (\mathcal{X}\cup \mathcal{Y})^{\star} \rightarrow \mathbb{R}^d$ and $\psi: (\mathcal{X}\cup \mathcal{Y})^{\star} \rightarrow \mathbb{R}^d$ are embedding functions\footnote{$\mathcal{A}^{\star}$ denotes the Kleene closure: the set of all finite concatenations of elements from $\mathcal{A}$.} used to encode the test sample $\mathbf{a}_{\text{test}}$ and in-context examples $\mathbf{a}_i$, respectively. The relevance between the test sample and candidate examples is typically scored using a similarity function—most often the dot product: $\phi(\mathbf{a}_{\text{test}})^{\top} \psi(\mathbf{a}_i)$.

\section{Proposed Framework}
\label{sec:methods}

\noindent\textbf{Problem Setup.} Reward models are central to alignment, scoring responses for coherence, relevance, safety, and other quality dimensions~\citep{bai2022training}. Yet, nearly all open-source reward models are trained on preference data drawn from high-resource languages~\citep{lambert2024rewardbench}. This mismatch can lead to inaccurate reward signals in low-resource contexts, limiting the effectiveness of these models in multilingual alignment~\citep{gureja2024m, she2024mapo}. Additionally, the scarcity of preference data in low-resource languages makes large-scale preference-based training of reward models impractical. This gap in capability raises a natural question:
\emph{Can we exploit reward models trained on high-resource languages to accurately judge response quality in low-resource Indic languages without expensive fine-tuning?}

We address this challenge by introducing a novel in-context learning strategy, \name, aimed at enhancing reward model performance in low-resource Indic languages.

\subsection{Our Approach: \name}

\name consists of a two-step framework that: (1) Identifies closely related high-resource Indic languages and leverages associated example banks to provide better context~\citep{ghosal2024promptrefine} (Section~\ref{subsubsec:aux_data_sel}), and (2) Refines retriever embeddings using pair-wise ranking loss to effectively select in-context examples with better discriminative context (Section~\ref{subsubsec:train_ret}).

\subsubsection{Auxiliary Dataset Selection}
\label{subsubsec:aux_data_sel}

Directly using few-shot examples from low-resource Indic languages often fails to improve model generalization due to limited and unrepresentative data~\citep{ghosal2024promptrefine}. To address this, we leverage closely related high-resource Indic languages as auxiliary datasets. The rationale is that high-resource languages such as Hindi are well represented in the reward model’s pre-training corpus~\citep{dubey2024llama}, unlike low-resource languages like Bodo.

Following ~\citet{ghosal2024promptrefine}, for each low-resource language, we compute the cosine similarity between its text embeddings and those of candidate auxiliary languages. Languages with similarity scores above a threshold (Algorithm~\ref{algo:aux_select}, Appendix~\ref{app:algorithm}) are selected as auxiliaries. By retrieving and presenting in-context examples from these related high-resource languages, we enrich the reward model’s contextual guidance, enabling more reliable evaluation of qualities such as harmlessness, coherence, and helpfulness in low-resource settings.

\vspace{0.2cm}
\noindent\textbf{Example bank curation for training.} Standard preference-based training~\citep{ouyang2022training, christiano2017deep} typically assumes access to preference data of the form \((\*x, \*y^{+}, \*y^{-})\), where each triplet consists of a prompt \(\*x\), a preferred response \(\*y^{+}\), and a less-preferred response \(\*y^{-}\) for the same prompt. This setup requires annotators to generate multiple responses per prompt and provide comparative judgments, making data collection both labor-intensive and costly. For low-resource languages, constructing such preference datasets is often prohibitively expensive~\citep{zhong2024opportunities, cohere2024ailanguagegap}.

To address this challenge, we adopt a more flexible approach that uses unpaired data with binary quality labels, eliminating the need for costly paired rankings. Specifically, we represent each sample as a triplet: $\*t = (\*x, \*y, l)$, where $\*x$ is the query, $\*y$ is a response and $l\in\{+,-\}$ represents a binary label indicating whether 
$\*y$ is a preferred response for the prompt $\*x$ given a specific reward objective (e.g., helpfulness, harmlessness). This flexibility is particularly advantageous for low-resource languages, where paired preference data is often scarce. Additionally, our method can readily integrate existing datasets with minimal overhead, requiring only binary quality labeling. For brevity, we denote a query–response pair as $\*a=(\*x,\*y)$, so that $\*t=(\*a, l)$.

Following this strategy, we split the example bank of the low-resource target language $\mathcal{T}$ as $\mathcal{D}_{\mathcal{T}}=\{\mathcal{D}^+_{\mathcal{T}}, \mathcal{D}^-_{\mathcal{T}}\}$, where $\mathcal{D}^+_{\mathcal{T}} = \{(\*x_i, \*y_i, +)\}_{i=1}^{N}$ represents the set of positive responses (reward-aligned) and $\mathcal{D}^-_{\mathcal{T}} = \{(\*x_j', \*y_j', -)\}_{j=1}^{N'}$ represents the set of negative (not reward-aligned) responses. For the auxiliary example bank, using Algorithm~\ref{algo:aux_select}, we select a set of related high-resource languages  
\(\mathcal{H}=\{\mathcal{H}_1,\dots,\mathcal{H}_P\}\). We denote the auxiliary example banks as $\mathcal{D}^{\text{aux}}=\{\mathcal{D}_{\mathcal{H}_1}, \cdots, \mathcal{D}_{\mathcal{H}_P}\}$ such that  $\mathcal{D}_{\mathcal{H}_p} =  \left(\mathcal{D}^+_{\mathcal{H}_p}, \mathcal{D}^-_{\mathcal{H}_p}\right) $, where $\mathcal{D}^+_{\mathcal{H}_p}$ and $\mathcal{D}^-_{\mathcal{H}_p}$
represents the set of positive and negative responses for language $\mathcal{H}_p$.

\subsubsection{Retriever training using pair-wise loss}
\label{subsubsec:train_ret}

Our main goal is to learn a set of retrievers
\(\{\mathcal{R}_{p}\}_{p=1}^{P}\), one for each auxiliary high-resource
language \(\mathcal{H}_{p}\). During inference,
given a test sample
$\*a_{\text{test}} = (\*x_{\text{test}},\*y_{\text{test}}) \in \mathcal{D}^{\text{test}}_{\mathcal{T}} $ from the
low-resource target language, the retriever \(\mathcal{R}_{p}\) selects a set of \(K\) in-context exemplars from the corresponding
example bank \(\mathcal{D}_{\mathcal{H}_p}\). We implement each retriever \(\mathcal{R}_{p}\) as a dual encoder parameterized by language-specific encoders \((\phi_p, \psi_{p})\), where \(\phi_p(\cdot)\) encodes examples from the low-resource language \(\mathcal{T}\), and \(\psi_{p}(\cdot)\) encodes exemplars from the high-resource auxiliary corpus \(\mathcal{D}_{\mathcal{H}_p}\).

\vspace{0.2cm}
\noindent\textbf{Training Objective.} For fine-tuning each dense retriever $\mathcal{R}_p$, we optimize the retriever to select high-resource examples that are most useful as prompts~\citep{rubin2021learning}. Specifically, for each target low-resource language instance $\*t_{\mathcal{T}}=(\*x_{\mathcal{T}}, \*y_{\mathcal{T}}, l_{\mathcal{T}}) \in \mathcal{D}_{\mathcal{T}}$, we construct two candidate sets: positive candidates $\mathcal{A}^+_{\mathcal{H}_p}  =\{(\*a_j, +)\}_{j=1}^F \subset \mathcal{D}_{\mathcal{H}^+_p}$ and negative candidates $\mathcal{A}^-_{\mathcal{H}_p} =\{(\*a_j, -)\}_{j=1}^F \subset \mathcal{D}_{\mathcal{H}^-_p} $. These candidate sets are formed by retrieving $F=25$ nearest examples based on the embedding similarity $\phi_p(\*a_{\mathcal{T}})^\top\psi_p(\*a_j)$. We provide ablation studies on the impact of different candidate set sizes $F$ in Appendix~\ref{app:ablations}.

Next, to compute the pairwise loss, we form a set of paired positive and negative contexts $\mathcal{E}_{\mathcal{H}_p} = \mathcal{A}^+_{\mathcal{H}_p} \times \mathcal{A}^-_{\mathcal{H}_p}$, where the set product ($\times$ )
 creates all possible pairings between the selected positive and negative candidates. Each context pair $\*e \in \mathcal{E}_{\mathcal{H}_p}$ is then scored using the reward model to determine its utility as context for $\*t_{\mathcal{T}}$:
\begin{equation}
    s(\*e; \*t_{\mathcal{T}}) = \pi_{\text{RM}}([\*e, \*x_{\mathcal{T}}], \*y_{\mathcal{T}}).
\end{equation}
where $[\cdot,\cdot]$ denotes text concatenation.  The optimal example pair for the sample $\*t_{\mathcal{T}}$ is selected as: $\widetilde{\*e}=
\argmax_{\*e \in \mathcal{E}_{\mathcal{H}_p}} l_{\mathcal{T}} * s(\*e;\*t_{\mathcal{T}})$.

Finally, the retriever $\mathcal{R}_{p}$ is fine-tuned to rank $\widetilde{\*e}$ highly among all pairs in $\mathcal{E}_{\mathcal{H}_p}$ by minimizing the following negative log-likelihood loss:

\small
\begin{align}
    &\min_{\phi_p, \psi_p} \mathcal{L}_{\text{pair}}(\mathcal{D}_{\mathcal{T}}, \mathcal{D}_{\mathcal{H}_p}; \phi_p, \psi_p) = \underset{\*t_{\mathcal{T}} \sim \mathcal{D}_{\mathcal{T}}}{\mathbb{E}}[\ell(\*t_{\mathcal{T}}, \mathcal{E}_{\mathcal{H}_p})] \\
    &\ell(\*t_{\mathcal{T}}, \mathcal{E}_{\mathcal{H}_p}; \phi_p, \psi_p) = -\log \frac{e^{\phi_p(\*a_{\mathcal{T}})^\top \psi_p(\widetilde{\*e})}}{\underset{\*e \in \mathcal{E}_{\mathcal{H}_p}}{\sum} e^{\phi_p(\*a_{\mathcal{T}})^\top \psi_p(\*e)}}
\end{align}
\normalsize

The complete training algorithm is provided in Algorithm~\ref{algo:main} (Appendix~\ref{app:algorithm}).

\subsubsection{Inference}
\label{subsubsec:inference}

During inference, given a test instance $\*a_{\text{test}} = (\*x_{\text{test}},\*y_{\text{test}})$ from the target language $\mathcal{T}$, we leverage the trained retrievers to select in-context examples. Specifically, for each auxiliary high-resource language $H_p$, we use the corresponding trained retriever $\widehat{R}_p = \{\widehat{\phi}_p, \widehat{\psi}_p\}$ to identify $K$ relevant in-context example pairs. Following the training setup, we construct a paired auxiliary bank $\mathcal{E}_{\mathcal{H}_p} = \mathcal{D}^+_{\mathcal{H}_p} \times \mathcal{D}^-_{\mathcal{H}_p}$ and the top-K pairs $\{\*e_1, \dots, \*e_K \}$ are selected based on the similarity score: $\argmax_{\*e_k \in \mathcal{E}_{\mathcal{H}_p}} \widehat{\phi}_p(\*a_{\text{test}})^\top \widehat{\psi}_p(\*e_k)$.
This procedure is repeated for all $P$ auxiliary high-resource languages, yielding a total of $\mathcal{C} = P*K$ in-context example pairs. These selected pairs are then used to condition the reward model when evaluating the quality of the test instance. The template used for concatenating the selected in-context examples is detailed in Appendix~\ref{app:prompt_template}.

\newcolumntype{?}{!{\vrule width 1pt}}
\newcolumntype{a}{>{\columncolor{myblue}}c}
\begin{table*}[!ht]
      \centering
       
        \resizebox{\textwidth}{!}{%
        \begin{tabular}{cccc?cc?cc?cc}
        \toprule
       \multirow{2}{1.75cm}{\centering Finetuning-Based} & \multirow{2}{*}{Methods} & \multicolumn{2}{c?}{Bodo} & \multicolumn{2}{c?}{Santali} & \multicolumn{2}{c?}{Manipuri} &  \multicolumn{2}{c}{Odia}\\ 
       \cmidrule{3-10}

       & & LM-3.1-8B & LM-3.2-3B & LM-3.1-8B & LM-3.2-3B & LM-3.1-8B & LM-3.2-3B & LM-3.1-8B & LM-3.2-3B \\

       \midrule
     
        \xmark & Zero-shot & 54.32 & 53.18 & 43.76 & 51.90 & 47.88 & 53.65 & 62.53 & 52.27 \\
      \xmark & Random & 55.41 & 52.67 & 45.88 & 50.66 & 45.39 & 54.03 & 56.45 & 53.77 \\
        \xmark & BM25 & 48.22 & 56.09 & 43.94 & 56.82 & 48.33 & 55.47 & 66.94 & 63.38  \\
       \xmark & Top-K & 57.53 & 58.91 & 45.84 & 55.48 & 50.91 & 51.32 & 68.34 & 60.72 \\
       \midrule
      \cmark & EPR &  58.74 & 60.57 & 46.66 & 59.11 & 54.18 & 55.64 & 71.67 & 62.01   \\
       
     \rowcolor{myblue}  \cmark & \name~(Ours) & \textbf{64.29} & \textbf{62.73} & \textbf{67.92} & \textbf{65.48} & \textbf{62.84} & \textbf{59.91} & \textbf{77.67} & \textbf{69.93} \\
  
      & Absolute Gain $(\Delta)$ & 5.55 & 2.16 & 21.26 & 6.37 & 8.66 & 4.27 & 6.00 & 6.55 \\

      \bottomrule
    \end{tabular}%
        }
 \caption{\small \textbf{Evaluation on PKU-SafeRLHF~\citep{ji2024pku}.} We report accuracy results for queries and responses translated into Bodo, Manipuri, Santali, and Odia, using two reward models based on LLAMA-3.1-8B (LM-3.1-8B)~\citep{liu2024skywork} and LLAMA-3.2-3B (LM-3.2-3B)~\citep{yang2024regularizing}. Best results are marked in \textbf{bold}.}
\label{tab:pku}
\end{table*}

\newcolumntype{?}{!{\vrule width 1pt}}
\newcolumntype{a}{>{\columncolor{myblue}}c}
\begin{table*}[!t]
      \centering
       
        \resizebox{\textwidth}{!}{%
        \begin{tabular}{cccc?cc?cc?cc}
        \toprule
       \multirow{2}{1.75cm}{\centering Finetuning-Based} & \multirow{2}{*}{Methods} & \multicolumn{2}{c?}{Bodo} & \multicolumn{2}{c?}{Santali} & \multicolumn{2}{c?}{Manipuri} &  \multicolumn{2}{c}{Odia}\\ 
       \cmidrule{3-10}

       & & LM-3.1-8B & LM-3.2-3B & LM-3.1-8B & LM-3.2-3B & LM-3.1-8B & LM-3.2-3B & LM-3.1-8B & LM-3.2-3B \\

       \midrule
     
        \xmark & Zero-shot & 50.32 & 51.17 & 61.00 & 56.43 & 59.26 & 51.08 & 50.07 & 56.23  \\
      \xmark & Random & 53.15 & 52.23 & 58.31 & 53.41 & 56.81 & 51.65 & 49.29 & 52.66  \\
        \xmark & BM25 &  60.12 & 58.94 & 59.48 & 57.37 & 56.03 & 54.43 & 61.01 & 60.15  \\
       \xmark & Top-K & 56.47 & 60.13 & 54.66 & 57.02 & 57.82 & 53.65 & 62.04 & 60.87  \\
       \midrule
      \cmark & EPR & 54.82 & 60.50 & 59.00 & 59.11 & 55.48 & 57.18 & 57.83 & 61.00   \\
       
     \rowcolor{myblue}  \cmark & \name~(Ours) & \textbf{62.91} & \textbf{65.88} & \textbf{66.29} & \textbf{67.42} & \textbf{68.33} & \textbf{63.49} & \textbf{67.43} & \textbf{66.12} \\
    
     & Absolute Gain $(\Delta)$ &  2.79 & 5.38 & 4.91 & 8.31 & 8.85 & 6.31 & 5.39 & 5.25 \\

      \bottomrule
    \end{tabular}%
        }
 \caption{\small \textbf{Evaluation on HH-RLHF~\citep{bai2022training}.} We report accuracy results for queries and responses translated into Bodo, Manipuri, Santali, and Odia, using two reward models. Best results are marked in \textbf{bold}.}
\label{tab:hh}
\end{table*}

\section{Experiments}
\label{sec:experiments}

In this study, we assess the generalization capabilities of two recent open-source classifier-based reward models from RewardBench~\citep{lambert2024rewardbench}, built on the LLAMA-3.1-8B~\citep{liu2024skywork}\footnote{Skywork/Skywork-Reward-Llama-3.1-8B-v0.2} and LLAMA-3.2-3B~\citep{yang2024regularizing}\footnote{Ray2333/GRM-Llama3.2-3B-rewardmodel-ft} architectures. We selected these models for their strong performance on high-resource Indic languages such as Hindi. Our experiments focus on the following Indic languages~\citep{singh2024indicgenbench}:

\begin{itemize}[left=0em]
\item \textbf{Low/Mid-Resource Languages}: Bodo, Santali, Manipuri, and Odia.
\item \textbf{Auxiliary High-Resource Languages}: Bengali, Hindi, Marathi, Gujarati, Kannada, Malayalam, Tamil, Telugu, and Urdu.
\end{itemize}

\noindent\textbf{Datasets.} A key challenge in evaluating reward models for low-resource Indic languages is the scarcity of high-quality preference datasets. To address this, we created multilingual versions of three widely used English-based preference datasets: PKU-SafeRLHF~\citep{ji2024pku}, HH-RLHF~\citep{bai2022training}, and WebGPT~\citep{nakano2021webgpt}. Each dataset was translated into all the listed Indic languages and manually curated to ensure both linguistic accuracy and contextual relevance. Detailed information on the dataset curation process is provided in Appendix~\ref{app:dataset_curation}.

\vspace{0.2cm}
\noindent\textbf{Baselines.} For a comprehensive evaluation, we compare \name with both fine-tuning-free retrievers such as Random, BM25~\cite{robertson2009probabilistic}, Top-K, and fine-tuned retrievers such as EPR~\citep{rubin2021learning}. We refer the reader to Appendix~\ref{app:baseline} for a detailed description of the baselines.

\vspace{0.2cm}
\noindent\textbf{Implementation Details.} In our proposed approach, \name, the retriever embeddings are initialized using a pre-trained multilingual BERT encoder\footnote{google-bert/bert-base-multilingual-cased}. We use an Adam optimizer with a batch size of $64$, a learning rate of $1e-4$, and fine-tune the retriever for $120$ epochs. All experiments are run using the configurations listed in Appendix~\ref{app:specs}. For all baselines, we fix the number of in-context examples
as $\mathcal{C}=8$ during inference, and we truncate it based on the maximum
context size of the reward model. Further, we also sort the in-context examples in ascending order of their similarity to the input query.

\vspace{0.2cm}
\noindent\textbf{Evaluation Criteria.} To systematically evaluate the quality of the in-context examples retrieved by \name, we measure pairwise accuracy on a held-out test set. Specifically, given a test query and pair of responses in target language $(\*x, \*y^+, \*y^-) \in \mathcal{D}^{\text{test}}_{\mathcal{T}}$, a reward evaluation is considered correct if the model assigns a higher score to the preferred response $(\*y^+)$ than to a less-preferred response $(\*y^-)$. 
As outlined in Section~\ref{subsubsec:inference}, for both the pairs $\*a^+=(\*x, \*y^+)$ and $\*a^-=(\*x, \*y^-)$, we independently select $\mathcal{C}$ in-context examples using \name. Let the set of retrieved examples for the preferred pair be $\{\*e^+_i\}_{i=1}^{\mathcal{C}}$ and for the less-preferred pair be $\{\*e^-_i\}_{i=1}^{\mathcal{C}}$ respectively. These examples are concatenated with the original query to form the augmented prompt: $\*x^+_{\text{aug}} = [\*e^+_1, \*e^+_2, \cdots, \*e^+_{\mathcal{C}}, \*x]$ and $\*x^-_{\text{aug}} = [\*e^-_1, \*e^-_2, \cdots, \*e^-_{\mathcal{C}}, \*x]$. The overall accuracy is then defined as:

\small
\begin{equation}
   \underset{(\*x, \*y^+, \*y^-) \sim \mathcal{D}^{\text{test}}_{\mathcal{T}}}{\mathbb{E}}\left[\mathbb{I}\left\{\pi_{\text{RM}}(\*x^+_{\text{aug}}, \*y^+) > \pi_{\text{RM}}(\*x^-_{\text{aug}}, \*y^-)\right\}\right],
\end{equation}
\normalsize
where \(\pi_{\text{RM}}\) represents the reward model, and \(\mathbb{I}\) is the indicator function. A higher accuracy indicates that the retrieved in-context examples help the reward model better distinguish preferred responses in low-resource target languages.

\newcolumntype{?}{!{\vrule width 1pt}}
\newcolumntype{a}{>{\columncolor{myblue}}c}
\begin{table*}[!t]
      \centering
       
        \resizebox{\textwidth}{!}{%
        \begin{tabular}{cccc?cc?cc?cc}
        \toprule
        \multirow{2}{1.75cm}{\centering Finetuning-Based} & \multirow{2}{*}{Methods} & \multicolumn{2}{c?}{Bodo} & \multicolumn{2}{c?}{Santali} & \multicolumn{2}{c?}{Manipuri} &  \multicolumn{2}{c}{Odia}\\ 
       \cmidrule{3-10}

        & & LM-3.1-8B & LM-3.2-3B & LM-3.1-8B & LM-3.2-3B & LM-3.1-8B & LM-3.2-3B & LM-3.1-8B & LM-3.2-3B \\

       \midrule
     
       \xmark & Zero-shot &  51.73 & 54.61 & 45.88 & 53.27 & 48.93 & 46.51 & 53.21 & 54.81 \\
       \xmark & Random & 50.86 & 50.34 & 46.77 & 50.67 & 47.83 & 51.62 & 51.97 & 49.73    \\
        \xmark & BM25 & 51.92 & 51.56 & 48.89 & 50.46 & 53.91 & 53.74 & 54.95 & 57.68  \\
        \xmark & Top-K & 54.98 & 54.42 & 53.92 & 48.91 & 45.91 & 49.81 & 54.11 & 53.93 \\
         \midrule
        \cmark & EPR & 50.62 & 57.29 & 48.53 & 50.78 & 55.45 & 50.39 & 55.42 & 57.81  \\
     
     \rowcolor{myblue}  \cmark & \name~(Ours) & \textbf{57.89} & \textbf{67.42} & \textbf{60.34} & \textbf{61.65} & \textbf{62.49} & \textbf{59.74} & \textbf{65.93} & \textbf{65.32} \\
    
      & Absolute Gain $(\Delta)$ & 2.91 & 10.13 & 6.42 & 8.38 & 6.58 & 5.99 & 10.98 & 7.64 \\

      \bottomrule
    \end{tabular}%
        }
 \caption{\small \textbf{Evaluation on WebGPT~\citep{nakano2021webgpt}.} We report accuracy results for queries and responses translated into Bodo, Manipuri, Santali, and Odia, using two reward models. Best results are marked in \textbf{bold}.}
\label{tab:webgpt}
\end{table*}

\subsection{Main Results}
\label{subsec:main_results}

We present our evaluation results on PKU-SafeRLHF, HH-RLHF, and WebGPT in Tables~\ref{tab:pku}, \ref{tab:hh}, and \ref{tab:webgpt}, respectively. Our observations are: 
\begin{enumerate}[itemsep=0.5pt, parsep=0.5pt, left=0cm]
    \item Across most languages and datasets, the zero-shot accuracy of reward models is close to random, typically around $50\%$. For certain languages, such as Santali and Manipuri, the reward model accuracy even drops below $50\%$, underscoring the severity of the challenge.

    \item Incorporating in-context examples using frozen pre-trained retrievers—whether through random sampling or Top-K selection—often fails to improve accuracy and can even lead to performance degradation. For instance, on the HH-RLHF dataset, Top-K retrieval for Santali reduces accuracy from $61.00\%$ to $54.66\%$. While fine-tuning the retriever with EPR~\citep{rubin2021learning} yields some accuracy gains over non-finetuned methods, these improvements are minimal. On PKU-SafeRLHF~\citep{ji2024pku} with the LLAMA-3.1-8B reward model, the maximum improvement achieved by EPR is only $3.27\%$. These results suggest that simply fine-tuning the retriever for relevance is insufficient to substantially enhance reward model performance.

    \item Retrieving in-context examples using \name provides consistent gains in accuracy over other example selection approaches.  For instance, on WebGPT, \name improves accuracy by $10.13\%$ for Bodo and $10.98\%$ for Odia compared to EPR~\citep{rubin2021learning}. Further, for Santali on PKU-SafeRLHF, \name achieves a substantial $21.26\%$ improvement over EPR. These results underscore the effectiveness of pairwise ranking loss in training the retriever, enabling it to provide more discriminative context and significantly enhance reward model generalization to low-resource languages.
\end{enumerate}

\section{Discussion}
\label{sec:discussion}

\newcolumntype{?}{!{\vrule width 1pt}}
\newcolumntype{a}{>{\columncolor{myblue}}c}
\begin{table}[!t]
      \centering
       
        \resizebox{\columnwidth}{!}{%
        \begin{tabular}{ccccccc}
        \toprule
    \makecell{Pair-wise \\ Loss} &    \makecell{Auxiliary\\High-resource Data}  & {Bodo} & {Santali} & {Manipuri} & {Odia} & Avg.\\ 
       \midrule
     
  \xmark & \xmark &  58.74 & 46.66 & 54.18 & 71.67 & 57.82\\
    \rowcolor{myblue} \cmark & \xmark  & 61.45 & 60.73 & 59.82 & 75.13 & 64.28\\
      \midrule
   \xmark &    \cmark &  62.36 & 58.36 &  60.99 & 74.84 & 64.12\\
   \rowcolor{myblue}   \cmark & \cmark  & 64.29 & 67.92 & 62.48 & 77.67 & 68.09 \\
      \bottomrule
    \end{tabular}%
        }
 \caption{\small We conduct ablation studies to assess the impact of using pairwise loss during retriever fine-tuning and the benefit of incorporating an auxiliary example bank. The dataset is PKU-SafeRLHF, and the reward model is based on LLAMA-3.1-8B.}
 \vspace{-4mm}
\label{tab:ablation}
\end{table}

\noindent\textbf{Understanding the importance of \name’s components.}
As evident from Section~\ref{subsec:main_results}, retrieving few-shot examples with \name leads to substantial improvements in accuracy. In this section, we further investigate the source of these gains through ablation studies on the key components of \name, conducted on the PKU-SafeRLHF dataset using the LLAMA-3.1-8B reward model.

\begin{enumerate}[itemsep=0pt, parsep=0pt, left=0pt]
\item \textbf{Impact of Pair-wise Ranking Loss.}
To evaluate the contribution of the pair-wise ranking loss, we fine-tune two retrievers for comparison: one optimized only for relevance~\citep{rubin2021learning}, and another trained with the pair-wise loss. Both models use the same auxiliary dataset configuration to ensure a fair comparison. As shown in Table~\ref{tab:ablation}, when auxiliary high-resource data is unavailable, the retriever fine-tuned with pairwise loss improves reward model accuracy by $6.46\%$ over the relevance-based baseline. When auxiliary data is available, the pairwise loss yields a $3.97\%$ improvement. These results underscore the value of pairwise ranking loss in enabling the retriever to select more effective and discriminative in-context examples.

\item \textbf{Importance of auxiliary high-resource example bank.} To examine the effect of auxiliary high-resource example banks, we train retrievers both with and without access to these example banks, using both relevance-based and pairwise loss for fine-tuning. From Table~\ref{tab:ablation} we observe that, regardless of the fine-tuning loss, access to auxiliary high-resource examples consistently improves performance.
\end{enumerate}

\begin{figure}[!t]
    \centering
    \includegraphics[width=0.88\columnwidth]{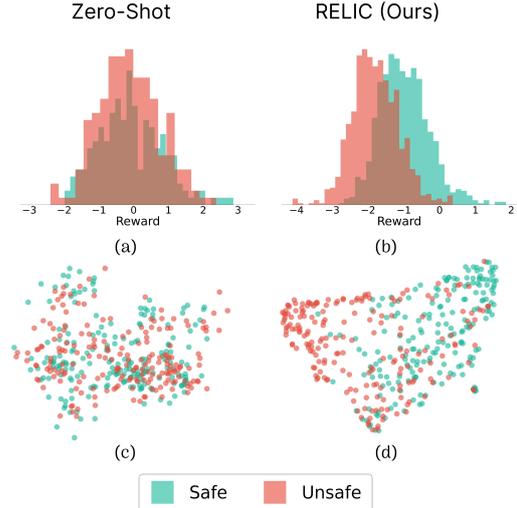}
    \caption{\small
\textbf{Top row:} Histogram of zero-shot reward scores for safe and unsafe responses in Santali shows substantial overlap, while incorporating in-context learning (ICL) examples from \name\ leads to more distinct and separable distributions.
\textbf{Bottom row:} UMAP visualization of the final hidden layer representations of the reward model for safe and unsafe responses in Santali, both with and without in-context examples selected by \name. }
\vspace{-4mm}
    \label{fig:umap_plot}
\end{figure}

\noindent\textbf{Why does \name improve reward model performance?} Previously, we highlighted how each component of \name contributes to improved performance. To better understand why few-shot samples from \name enhance reward model accuracy on low-resource languages, we analyze the final hidden layer representations of the reward model. In Figure~\ref{fig:umap_plot}, we compare how the reward model represents safe and unsafe responses in Santali, both with and without in-context examples selected by \name, using queries from HH-RLHF. Our observations are as follows: (1) For zero-shot prompting (Figure~\ref{fig:umap_plot}(c)), the representations of safe and unsafe responses are highly entangled, which explains the poor zero-shot performance. (2) When in-context examples from \name are incorporated, the representations of safe and unsafe responses become much more distinct (Figure~\ref{fig:umap_plot}(d)), resulting in higher accuracy. We extend this analysis to additional languages in Figure~\ref{fig:additional_umap} (Appendix).

\section{Conclusion}
\label{sec:conclusion}

In this paper, we analyze the generalization capabilities of open-source multilingual reward models on low-resource Indic languages. Our results show that, despite strong performance on high-resource languages, these models often fail to provide reliable reward signals in low-resource contexts. To address this, we introduce \name, a few-shot learning approach that fine-tunes a retriever with a pairwise ranking loss to select in-context examples that are both relevant and discriminative. We further enhance performance by leveraging auxiliary example banks from related high-resource languages. Extensive experiments across multiple preference datasets, languages, and state-of-the-art open-source reward models demonstrate that in-context examples selected by \name consistently yield more reliable reward signals in low-resource settings—without the need for costly fine-tuning.

\section{Limitations and Future work}

While \name demonstrates promising results in enhancing reward model generalization, it has the following limitations:

\begin{itemize}
\item Our evaluation of \name is limited to a subset of low-resource Indic languages. Although the method itself is language-agnostic, its effectiveness on other language families remains to be validated.

\item This study primarily focuses on reward model accuracy as the evaluation metric. The impact of incorporating in-context examples on multilingual alignment strategies is a potential direction for future research.

\item Our analysis focuses on classifier-based reward models trained with the Bradley-Terry framework~\citep{bradley1952rankanalysis, christiano2017deep, ouyang2022training}. Exploring how \name\ could be extended to implicit or generative reward models is an interesting avenue for future work.
\end{itemize}

\clearpage
\newpage

\bibliography{custom_cleaned}

\clearpage
\newpage
\appendix

\section{Software and Hardware}
\label{app:specs}
We run all experiments with Python 3.12.8 and PyTorch 2.6.0. For all experimentation, we use one Nvidia RTX A6000 GPU. 

\section{Experimental Details}
\label{app:experiment_details}

\subsection{Dataset Curation}
\label{app:dataset_curation}

Obtaining high-quality preference test sets in low-resource Indic languages is challenging. To address this, we curated our test set by translating three widely used English-based preference datasets—PKU-SafeRLHF~\citep{ji2024pku} (\texttt{PKU-Alignment/PKU-SafeRLHF}), HH-RLHF~\citep{bai2022training} (\texttt{Dahoas/full-hh-rlhf}), and WebGPT~\citep{nakano2021webgpt} (\texttt{openai/webgpt\_comparisons})—into all Indic languages considered in Section~\ref{sec:experiments}. Translations were generated using the \texttt{ai4bharat/indictrans2-indic-en-1B} model and manually reviewed to ensure linguistic accuracy.

For all languages and preference datasets, we set the size of the target low-resource language example bank $\mathcal{D}_{\mathcal{T}}$ to 1{,}000 and the test set $\mathcal{D}^{\text{test}}_{\mathcal{T}}$ to 700. For all auxiliary high-resource languages $D_{\mathcal{H}_p}$, the example bank size is set to 5{,}000.

\subsection{Baseline Details}
\label{app:baseline}

For a comprehensive evaluation, we compare \name\ against the following baselines:

\begin{itemize}[noitemsep, leftmargin=*]
\item \textbf{Random:} In-context examples are randomly sampled from the training set without replacement.
\item \textbf{BM25:} A classical sparse retrieval method, BM25~\cite{robertson2009probabilistic}, is used to rank and select the highest-scoring examples as in-context samples.
\item \textbf{Top-K:} This baseline employs a dense retriever initialized with pre-trained multilingual BERT embeddings, without any fine-tuning. The top-ranked examples are chosen as in-context samples.
\item \textbf{EPR}~\citep{rubin2021learning}: The multilingual BERT retriever is fine-tuned using a relevance loss; at inference, the highest-ranked examples are used as in-context samples.
\end{itemize}

\section{Extended Discussion}
\label{app:ablations}

\begin{figure}[!t]
    \centering
    \includegraphics[width=\linewidth]{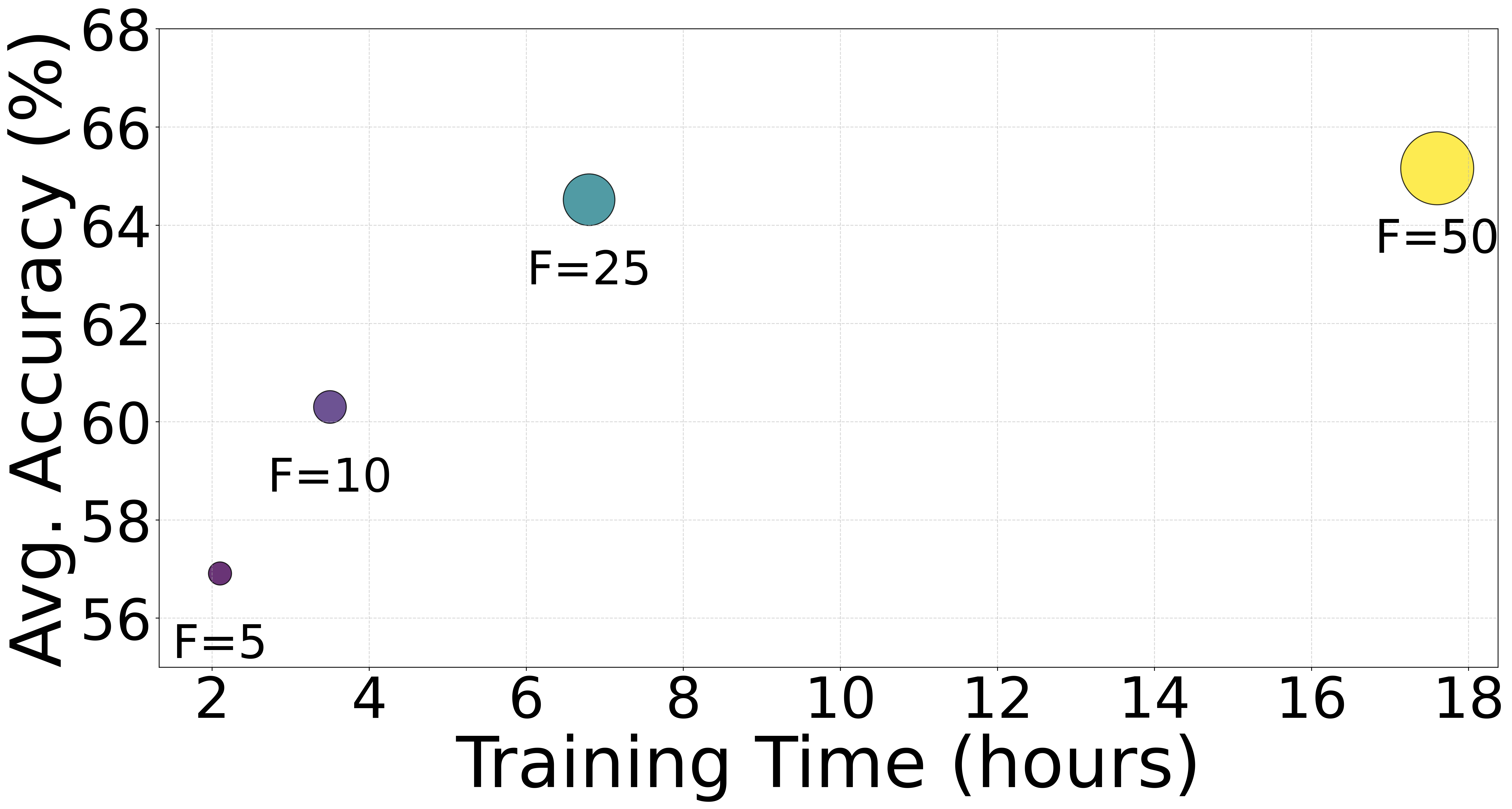}
    \caption{We visualize the effect of varying the candidate set size $F$ on both accuracy and training time. The reward model used is LLAMA-3.1-8B, with accuracy averaged over Bodo, Santali, Manipuri, and Odia queries from the PKU-SafeRLHF dataset.}
    \label{fig:ablation_f}
\end{figure}

\paragraph{Ablation study on candidate set size $F$.}
When training the retriever with pairwise ranking loss, we construct both the positive and negative candidate sets with size $F$. In Figure~\ref{fig:ablation_f}, we visualize the effect of varying the candidate set size $F \in \{5, 10, 25, 50\}$ on overall performance and training time. The reward model used is LLAMA-3.1-8B, and accuracy is averaged over Bodo, Santali, Manipuri, and Odia queries from PKU-SafeRLHF.

We observe that while a smaller candidate set ($F=5$) results in the lowest training time, it comes at the cost of reduced accuracy. Conversely, a larger candidate set ($F=50$) yields the highest accuracy but introduces significant computational overhead. Based on this trade-off, we select $F=25$ for all our experiments.

\begin{figure}[!t]
    \centering
    \includegraphics[width=\linewidth]{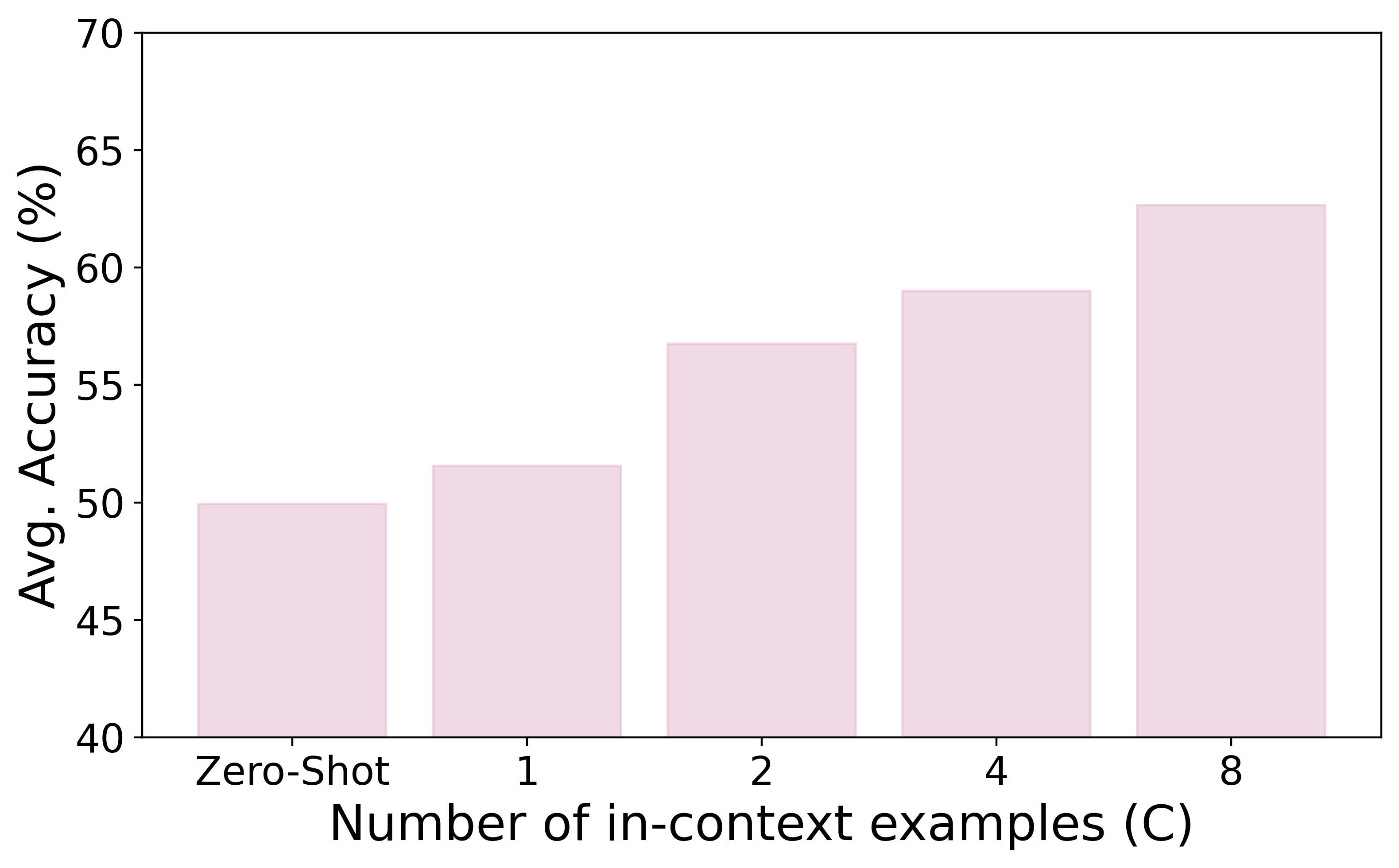}
    \caption{We visualize the effect of the number of in-context examples on reward model accuracy.}
    \label{fig:ablation_c}
\end{figure}

\paragraph{Effect of the number of in-context examples $\mathcal{C}$.}
In Figure~\ref{fig:ablation_c}, we present an ablation study on the number of in-context examples, $\mathcal{C}$. We evaluate performance for $\mathcal{C} = \{1, 2, 4, 8\}$ on the HH-RLHF dataset across all low-resource languages. Due to computational constraints and the limited context window of the reward model, we cap the maximum number of in-context examples at $8$. Our results show a monotonic increase in reward model accuracy as the number of in-context examples increases, with $\mathcal{C} = 8$ achieving the best performance in this setting.


\section{Algorithms}
\label{app:algorithm}

We present the training procedure for \name\ in Algorithm~\ref{algo:main}, and the method for selecting closely related high-resource example banks in Algorithm~\ref{algo:aux_select}. Additionally, consistent with \citet{ghosal2024promptrefine}, we set the threshold parameter $\gamma$ in Algorithm~\ref{algo:aux_select} to the 95th percentile of cosine similarity values between target low-resource language embeddings and those of high-resource languages.

\definecolor{systemcol}{RGB}{220,20,60}   
\definecolor{usercol}{RGB}{34,139,34}     
\definecolor{assistcol}{RGB}{30,144,255}  
\definecolor{icecol}{RGB}{138,43,226}     
\definecolor{langcol}{RGB}{255,165,0}     

\section{Prompt Template}
\label{app:prompt_template}

We describe the prompt template used for evaluating the reward model in Table~\ref{tab:prompt-template}.
\newcolumntype{?}{!{\vrule width 1pt}}

\begin{table*}[t]
  \centering
  \resizebox{\textwidth}{!}{%
    \begin{tabular}{@{}p{0.18\textwidth}p{0.78\textwidth}@{}}
      \toprule
      \textbf{Scenario} & \textbf{Prompt Structure} \\
      \midrule
      \textbf{Zero-shot} &
        \textcolor{systemcol}{\textbf{System:}} You are an accurate reward model. Your goal is to evaluate a (Query, Response) pair and output a numerical reward score that reflects how well the response answers the query. A higher score means the response is of higher quality, safer, and relevant. A lower score means the response is incorrect, irrelevant, unsafe, or otherwise poor.\\
      & \textcolor{usercol}{\textbf{User:}} \emph{(Query)}\\
      & \textcolor{assistcol}{\textbf{Assistant:}} \emph{(Response)}\\
      \addlinespace
      \textbf{\name } &
        \textcolor{systemcol}{\textbf{System:}} The following are (Query, Response) examples in \textcolor{langcol}{[auxiliary language]}. Each is labeled as [positive response] or [negative response]. Positive and Negative examples appear in any order. The final example in \textcolor{langcol}{[target language]} language is for evaluation.\\
      & \quad \textbf{For \(i = 1,\dots, \mathcal{C}\):}\\
      & \quad\quad \textcolor{usercol}{User:} Query example \(i\) in the \textcolor{langcol}{auxiliary language}.\\
      & \quad\quad \textcolor{assistcol}{Assistant:} Response example \(i\), labeled [Positive response] or [Negative response].\\
      & \textcolor{usercol}{User:} Query in the \textcolor{langcol}{target language}.\\
      & \textcolor{assistcol}{Assistant:} Response in the \textcolor{langcol}{target language}.\\
      \bottomrule
    \end{tabular}%
  }
  \caption{Prompt templates used for evaluating reward‐models.}
  \label{tab:prompt-template}
\end{table*}

\begin{figure*}[t]
    \centering

    \begin{minipage}[b]{0.48\linewidth}
        \centering
        \includegraphics[width=\linewidth]{images/LM-3.1-8B-bodo.pdf}
        \caption*{(a) Bodo}
    \end{minipage}
    \hfill
    \begin{minipage}[b]{0.48\linewidth}
        \centering
        \includegraphics[width=\linewidth]{images/LM-3.1-8B-santali.pdf}
        \caption*{(b) Santali}
    \end{minipage}

    \vspace{0.5em} 

    \begin{minipage}[b]{0.48\linewidth}
        \centering
        \includegraphics[width=\linewidth]{images/LM-3.1-8B-Manipuri.pdf}
        \caption*{(c) Manipuri}
    \end{minipage}
    \hfill
    \begin{minipage}[b]{0.48\linewidth}
        \centering
        \includegraphics[width=\linewidth]{images/LM-3.1-8B-odia.pdf}
        \caption*{(d) Odia}
    \end{minipage}

    \caption{UMAP visualizations of the final hidden layer representations for queries and responses from PKU-SafeRLHF across different languages. The reward model is LLAMA-3.1-8B.}
    \label{fig:additional_umap}
\end{figure*}

\begin{algorithm*}[!h]
\caption{\name: Training the retriever}
\label{algo:main}
\begin{algorithmic}[1]
\State \textbf{Input:} Low-resource language example bank $\mathcal{D}_{\mathcal{T}}$; Auxiliary example banks $\mathcal{D}^{\text{aux}}$

\State $\Phi \gets \emptyset$
\For {each dataset $\mathcal{D}_p \in \{\mathcal{D}_{\mathcal{H}_1}, \cdots, \mathcal{D}_{\mathcal{H}_P}\}$}
    \State $\phi_p, \psi_p \gets \text{MBERT}$ 
\Comment{Initialize the retriever embedding with pre-trained Multi-lingual BERT encoder}
        \State $(\widehat{\phi_p}, \widehat{\psi}_p) \gets \argmin_{\phi_p, \psi_p} \mathcal{L}_{\text{pair}}(\mathcal{D}_{\mathcal{T}}, \mathcal{D}_p)$
        \State $\widehat{\mathcal{R}}_p = \{\widehat{\phi_p}, \widehat{\psi}_p\}$
        \State $\Phi \gets \Phi \cup \widehat{\mathcal{R}}_p$ \Comment{The set of all language-specific trained retrievers}
    \EndFor

\State \Return  $\Phi$
\end{algorithmic}
\end{algorithm*}

\begin{algorithm*}[t]
\caption{High-resource Indic language example bank selection}
\label{algo:aux_select}
\begin{algorithmic}[1]

\State \textbf{Input:} High-resource Indic language example banks  $\mathcal{D}^{\text{high}}=\{\mathcal{D}_{\mathcal{H}_1}, \cdots, \mathcal{D}_{\mathcal{H}_O}\}$; low-resource target language examples $\mathcal{D}_{\mathcal{T}}=\left\{ (\*x_{i}, \*y_{i}, l_i) \right\}_{i=1}^{N_{\mathcal{T}}}$; pre-trained multi-lingual BERT encoder $\rho$; Threshold $\gamma$.

\State $e_{\mathcal{T}} \gets \cfrac{1}{N_{\mathcal{T}}}\sum_{i=1}^{N_{\mathcal{T}}}\rho((\*x_{i}, \*y_{i}))$ 

\State $\text{sim} \gets \emptyset$

\For{each $\mathcal{D}_h \in \mathcal{D}^{\text{high}}$}
    \State $e_{h} \gets \cfrac{1}{N_h}\sum_{j=1}^{N_h}\rho((\*x_{jh}, \*y_{jh}))$ 
    \State $\text{sim}_{h} \gets \cfrac{e_{h}^{\top}e_{\mathcal{T}}}{|e_{h}||e_{\mathcal{T}}|}$

    \State $\text{sim} \gets \text{sim} \cup \text{sim}_{h}$

\EndFor

\State $\mathcal{D}^{\text{aux}} \gets \{\mathcal{D}_h \in \mathcal{D}^{\text{high}} \mid \text{sim}_h \geq \text{percentile}(\text{sim}, \gamma)\}$

\State \Return $\mathcal{D}^{\text{aux}}$ 
\end{algorithmic}
\end{algorithm*}

\end{document}